\pdfoutput=1
\documentclass[11pt]{article}

\usepackage[final]{acl}

\usepackage{times}
\usepackage{latexsym}
\usepackage{tcolorbox}
\usepackage{amsmath}
\usepackage{amssymb}
\usepackage[T1]{fontenc}
\usepackage{enumitem}
\usepackage{graphicx}
\usepackage{subcaption}
\usepackage{booktabs}
\usepackage{multirow}
\usepackage{adjustbox}

\usepackage[linesnumbered,ruled,vlined]{algorithm2e}

\usepackage[utf8]{inputenc}

\usepackage{microtype}
\usepackage{makecell} 
\usepackage{inconsolata}

\usepackage{graphicx}

\title{When Actions Teach You to Think: Reasoning-Action Synergy via Reinforcement Learning in Conversational Agents}

\author{
  Mrinal Rawat\thanks{Work done while time spent at Uniphore.} \\
  Uniphore
  \And
  Arkajyoti Chakraborty \\
   Uniphore \\
  \And 
  Neha Gupta \\
   Uniphore
  \And
  Roberto Pieraccini \\
   Uniphore
}

\begin{document}
\maketitle

\begin{abstract}
Supervised fine-tuning (SFT) has emerged as one of the most effective ways to improve the performance of large language models (LLMs) in downstream tasks. However, SFT can have difficulty generalizing when the underlying data distribution changes, even when the new data does not fall completely outside the training domain. Recent reasoning-focused models such as o1 and R1 have demonstrated consistent gains over their non-reasoning counterparts, highlighting the importance of reasoning for improved generalization and reliability. However, collecting high-quality reasoning traces for SFT remains challenging—annotations are costly, subjective, and difficult to scale. To address this limitation, we leverage Reinforcement Learning (RL) to enable models to learn reasoning strategies directly from task outcomes. We propose a pipeline in which LLMs generate reasoning steps that guide both the invocation of tools (e.g., function calls) and the final answer generation for conversational agents. Our method employs Group Relative Policy Optimization (GRPO) with rewards designed around tool accuracy and answer correctness, allowing the model to iteratively refine its reasoning and actions. Experimental results demonstrate that our approach improves both the quality of reasoning and the precision of tool invocations, achieving a 1.5\% relative improvement over the SFT model (trained without explicit thinking) and a 40\% gain compared to the base of the vanilla Qwen3-1.7B model. These findings demonstrate the promise of unifying reasoning and action learning through RL to build more capable and generalizable conversational agents.
\end{abstract}

\section{Introduction}

Recent advances in large language models (LLMs) have demonstrated remarkable capabilities in multi-step reasoning, planning, and decision-making, extending beyond simple text generation. Early methods such as chain-of-thought (CoT) prompting \cite{DBLP:journals/corr/abs-2201-11903} revealed that LLMs can produce intermediate reasoning steps, improving performance in arithmetic, common sense, and logical reasoning tasks. Subsequent techniques, including self-consistency \cite{DBLP:conf/iclr/0002WSLCNCZ23}, step-by-step verification \cite{lightman2023letsverifystepstep}, and Tree-of-Thoughts \cite{10.5555/3666122.3666639}, further enhanced reasoning by exploring multiple solution paths and verifying intermediate results. Approaches such as ReAct \cite{yao2022react} and planning-based methods \cite{rawat2025preactmultistepplanningreasoning} integrate reasoning with tool use using supervised fine-tuning (SFT), allowing models to solve complex tasks.

Although SFT can drastically improve task performance compared to prompt-based vanilla methods, it typically requires large, high-quality datasets. Recent evidence from reasoning-focused models like o1 and R1 \cite{deepseekai2025deepseekr1incentivizingreasoningcapability} demonstrates that models exhibiting explicit reasoning consistently outperform their non-reasoning counterparts and generalize better across domains. Moreover, standard SFT approaches that omit reasoning often fail to develop emergent capabilities and struggle with out-of-distribution tasks. Obtaining high-quality reasoning annotations presents significant challenges: the process is expensive, subjective, and prone to biases that can lead to overthinking, incoherent, or even contradictory reasoning traces that ultimately hurt performance.

Reinforcement learning (RL) offers a compelling alternative to supervised annotation. Rather than requiring expert-labeled reasoning traces, RL enables models to discover effective reasoning strategies through interaction and feedback from task outcomes. This approach has proven to be  transformative: DeepSeek-R1 demonstrated that RL can elicit sophisticated reasoning behaviors without explicit reasoning supervision, while ToolRL \cite{qian2025toolrlrewardtoollearning} showed that reward-driven learning significantly improves tool-use accuracy in agentic systems. The key insight is that by optimizing task success through trial and error, models can learn not just \emph{what} actions to take, but \emph{how to reason} about when and why to take them, creating a synergistic relationship between reasoning and action.

In this work, we propose a methodology that helps models generate effective reasoning, which in turn improves their performance on actions such as tool calling. We leverage Group Relative Policy Optimization (GRPO) to jointly optimize reasoning and action generation, using various set of rewards derived from the precision of action and the correctness of the responses. Our approach requires minimal reasoning annotations while achieving strong performance through reward-driven discovery of effective reasoning patterns. Our contributions are:
\begin{itemize}
    \item We propose a three-stage pipeline: (i) \textbf{Base SFT} on task data without reasoning to establish instruction-following capabilities; (ii) \textbf{Cold-start Reasoning SFT} using a small reasoning-annotated dataset (<100 samples) to initialize structured reasoning; and (iii) \textbf{Reinforcement Learning with Verifiable Rewards} that jointly optimizes reasoning and actions via GRPO\cite{shao2024deepseekmathpushinglimitsmathematical}.
    \item We design composite rewards that balance conditional decision-making (when to call tools vs. answer directly), tool correctness, reasoning efficiency, output format adherence, and answer accuracy, enabling simultaneous improvement in reasoning and conversational task success.
    \item We conduct experiments on tool-augmented conversational benchmarks and present results at each stage, demonstrating that our approach of post-training with verifiable rewards outperforms both the vanilla and supervised models.
\end{itemize}


\section{Related Works}

The emergence of large language models has fundamentally transformed conversational AI, enabling sophisticated multi-turn interactions and seamless tool integration~\cite{gao2023largelanguagemodelsempowered, li2025chatsopsopguidedmctsplanning}. Earlier systems relied on rule-based or statistical methods with limited adaptability, whereas LLM-powered agents now have the power to achieve near-human performance in customer service, personal assistance, and complex recommendation tasks with minimal intervention from conversation designers and software engineers~\cite{ma2023understandingbenefitschallengesusing, becker2024multiagentlargelanguagemodels}. Planning-based approaches enhance controllability and proactivity in goal-oriented dialogues, and multi-agent frameworks distribute sub-tasks among specialized models to improve coherence and scalability~\cite{li2025chatsopsopguidedmctsplanning, rawat2025preactmultistepplanningreasoning}. Despite these advances, maintaining task adherence and avoiding hallucinations remain open challenges, especially in extended or safety-critical conversations~\cite{becker2024multiagentlargelanguagemodels}.

Supervised fine-tuning (SFT) has been the dominant paradigm for adapting pretrained LLMs, but its reliance on large annotated corpora and propensity to overfit limit its reasoning and generalization capabilities~\cite{pareja2024unveilingsecretrecipeguide,jin2025rlpanaceamirageunderstanding}. Reinforcement learning (RL) addresses these gaps by optimizing through interaction and outcome-based feedback, reducing annotation needs while enhancing robustness to out-of-distribution inputs~\cite{ye2025robustreinforcementlearninghuman}. Additionally, RL improves the models not only by reinforcing correct behavior, but also by reducing the frequency of incorrect or undesired outcomes. DeepSeek-R1 demonstrated that pure RL training without extensive SFT can elicit emergent chain-of-thought and self-correction behaviors in LLMs~\cite{deepseekai2025deepseekr1incentivizingreasoningcapability}. Hybrid methods like SRFT integrate SFT stability with RL exploration to balance convergence speed and reasoning quality~\cite{fu2025srftsinglestagemethodsupervised}.

Applying RL to conversational agents re-frames dialogue as a sequential decision problem, where agents learn policies for choosing responses, questions, or tool calls to maximize long-term conversational success. Tool-calling has emerged as a critical capability, with frameworks like ARTIST \cite{singh2025agenticreasoningtoolintegration} and ToolRL \cite{qian2025toolrlrewardtoollearning} using RL to learn when and how to invoke external APIs based solely on task rewards. ReCall further showed that complex tool-based reasoning can be acquired without supervised tool-use trajectories~\cite{chen2025research}. Recent surveys outline the rise of agentic RL, emphasizing the need for planning, memory, and autonomous decision-making in partially observable environments~\cite{zhang2025landscapeagenticreinforcementlearning}. Unlike Step-Wise RL (SWiRL)~\cite{goldie2025syntheticdatageneration}, which learns multi-step reasoning and tool use directly via RL with synthetic trajectories, we decompose the process into three phases. First, we leverage ground-truth action data and apply SFT to achieve substantial improvements in tool calling and response generation over the base pre-trained model. Next, a cold-start stage provides initial guidance for reasoning, and finally, we apply RL through GRPO~\cite{shao2024deepseekmathpushinglimitsmathematical}, we create a mutually reinforcing synergy that enhances both reasoning and task performance.

\section{Approach}
We adopt a cumulative learning strategy inspired by Pre-Act~\cite{rawat2025preactmultistepplanningreasoning}, progressively enhancing model capabilities from basic task execution to sophisticated reasoning-action coordination. Our training pipeline consists of three stages, each building upon the previous one to achieve robust performance.

\subsection{Three-Stage Training Pipeline}

\subsubsection{Base SFT: Establishing Task Foundations}
In the first stage, we fine-tune the pretrained model on annotated conversational data without reasoning traces. The goal is to develop foundational multi-turn conversational capabilities, including deciding whether to reply directly or invoke a tool, and, when invoking a tool, choosing the appropriate one with accurate parameters. Answer generation occurs in two scenarios: (1) direct responses when no tool is required, and (2) synthesized responses based on tool outputs. We employ Low-Rank Adaptation (LoRA)~\cite{hu2021loralowrankadaptationlarge} for parameter-efficient fine-tuning, preserving pre-trained knowledge while adapting to task-specific behaviors.

\subsubsection{Cold-Start Reasoning SFT: Initializing Structured Thinking}
The second stage introduces structured reasoning through fine-tuning on a small, high-quality reasoning-annotated dataset (<100 samples). This "cold-start" phase serves two critical purposes: (1) it bootstraps the model with reasoning patterns prior to reinforcement learning, and (2) it accelerates convergence during RL, as demonstrated in DeepSeek-R1. Training continues from the Base SFT checkpoint, ensuring retention of execution skills while introducing reasoning capabilities. The output format is standardized as:

\begin{tcolorbox}[colback=blue!5, colframe=blue!40, boxrule=0.8pt, arc=3mm, left=5pt, right=5pt, top=5pt, bottom=5pt]
\small
\texttt{<think>}\\
\quad\textit{[Reasoning]}\\
\texttt{</think>}\\[0.3em]
\texttt{<tool\_call>}\\
\quad\texttt{\{"name": "tool\_name", "arguments": \{...\}\}}\\
\texttt{</tool\_call>}\\[0.3em]
\textbf{OR}\\[0.3em]
\texttt{<answer>}\\
\quad\textit{Response to the user}\\
\texttt{</answer>}
\end{tcolorbox}

\subsubsection{Reinforcement Learning: Learning Reasoning Strategies}
In the final stage, we apply reinforcement learning using Group Relative Policy Optimization (GRPO) to jointly optimize reasoning quality and task performance. Starting from the cold-start checkpoint, the model learns to generate reasoning traces that maximize task success through reward-driven exploration.

\paragraph{GRPO Optimization.}
GRPO~\cite{shao2024deepseekmathpushinglimitsmathematical} extends PPO by eliminating the need for a critic model, instead using group-wise advantage estimation for variance reduction. For each question $q$, we sample $G$ outputs $\{o_1, \ldots, o_G\}$ from the old policy $\pi_{\theta_{\text{old}}}$ and optimize:
\begin{equation}
\begin{split}
\mathcal{L}_{\text{GRPO}}(\theta) = \mathbb{E} \Big[ \frac{1}{G} \sum_{i=1}^G \frac{1}{|o_i|} \sum_{t=1}^{|o_i|} \min \Big( r_t(\theta) \hat{A}_{i}, \\
\text{clip}(r_t(\theta), 1-\epsilon, 1+\epsilon) \hat{A}_{i} \Big) - \beta D_{\text{KL}}(\pi_\theta \| \pi_{\text{ref}}) \Big]
\end{split}
\end{equation}
where $r_t(\theta) = \frac{\pi_\theta(o_{i,t}|q, o_{i,<t})}{\pi_{\theta_{\text{old}}}(o_{i,t}|q, o_{i,<t})}$, and the group-normalized advantage is:
\begin{equation}
\hat{A}_{i} = \frac{r_i - \text{mean}(\mathbf{r})}{\text{std}(\mathbf{r})}, \quad \mathbf{r} = \{r_1, \ldots, r_G\}
\end{equation}

Here, $\pi_{\theta_{\text{old}}}$ is the sampling policy, $\pi_{\text{ref}}$ is the cold-start checkpoint, $\epsilon$ controls clipping, and $\beta$ regulates KL divergence.

\paragraph{Reinforcement Learning with Verifiable Rewards}
We design a composite reward function that balances decision-making accuracy, structural compliance, and reasoning adequacy:
\begin{equation}
\mathcal{R}_{\text{total}} = \mathcal{R}_{\text{cond}} + \mathcal{R}_{\text{fmt}} + \mathcal{R}_{\text{len}}
\end{equation}

\noindent\textbf{Conditional Accuracy Reward} ($\mathcal{R}_{\text{cond}} \in [-2, 2]$):
This reward integrates both action selection correctness and execution quality. Given ground truth $G$ and prediction $P$:
\begin{equation}
\mathcal{R}_{\text{cond}} =
\begin{cases}
1.0 + \frac{\mathcal{S}_{\text{tool}}}{3} & \text{if both use tool} \\[0.3em]
1.0 + \mathcal{S}_{\text{sem}} & \text{if both answer directly} \\[0.3em]
-2.0 & \text{otherwise}
\end{cases}
\end{equation}

The tool-matching score $\mathcal{S}_{\text{tool}} \in [-3, 3]$ evaluates three components, adapted from ToolRL~\cite{qian2025toolrlrewardtoollearning}:
\begin{equation}
\mathcal{S}_{\text{tool}} = 2 \cdot (s_{\text{name}} + s_{\text{keys}} + s_{\text{vals}}) - 3
\end{equation}
where:
\begin{itemize}
    \item $s_{\text{name}} = \begin{cases} 1 & \text{N}(G) = \text{N}(P) \\ 0 & \text{otherwise} \end{cases}$ where N represents the tool name
    \item $s_{\text{keys}} = \dfrac{|K_G \cap K_P|}{|K_G \cup K_P|} \in [0,1]$ measures the overlap of argument keys $K_G$ and $K_P$
    \item $s_{\text{vals}} = \dfrac{|{k \in K_G \cap K_P : G[k] = P[k]}|}{|K_G|} \in [0,1]$ computes the fraction of matching key-value pairs.
\end{itemize}

For answer, semantic similarity $\mathcal{S}_{\text{sem}} \in [0, 1]$ is computed using a cross-encoder re-ranker model \cite{xiao2024cpackpackedresourcesgeneral} that measures the semantic alignment between predicted and ground-truth answers, capturing correctness beyond exact string matching. This design heavily penalizes incorrect action selection ($-2.0$) while providing graded rewards for execution quality.

\noindent\textbf{Format Compliance Reward} ($\mathcal{R}_{\text{fmt}} \in \{0, 1\}$):
Ensures adherence to the required output structure. Let $C = \{\text{has\_think}, \text{has\_action}, \text{correct\_order}\}$ be the set of required format conditions. Then:
\begin{equation}
\mathcal{R}_{\text{fmt}} = \begin{cases}
1 & \text{if all conditions in } C \text{ are satisfied} \\
0 & \text{otherwise}
\end{cases}
\end{equation}
This requires the presence of both \texttt{<think>} and an action tag (\texttt{<tool\_call>} or \texttt{<answer>}) in the proper sequence (reasoning before action).

\noindent\textbf{Thinking Length Reward} ($\mathcal{R}_{\text{len}} \in \{0, 0.5, 1\}$):
Encourages sufficient yet concise reasoning:
\begin{equation}
\mathcal{R}_{\text{len}} =
\begin{cases}
1.0 & \text{if } |\text{tokens}| \in (m, n] \\[0.3em]
0.5 & \text{if } |\text{tokens}| > n \\[0.3em]
0.0 & \text{if } |\text{tokens}| \leq m
\end{cases}
\end{equation}
where $|\text{tokens}|$ denotes the word count within the \texttt{<think>} block. These values were chosen empirically based on the dataset. The lower bound of 14 tokens ensures that the reasoning can at least include the mention of the tool name. The upper bound of 100 tokens was selected because, in our cold-start data, all reasoning samples were under 80 tokens. We set it slightly higher to avoid excessive truncation. These settings strike a balance between reasoning depth and efficiency, helping the model learn when and how to act, thereby promoting effective coordination between reasoning and actions.

\section{Experiments}
\subsection{Datasets}

We conduct experiments on two multi-turn conversational agent datasets:

\noindent\textbf{APIGen-MT-5k} We use the Salesforce APIGen-MT-5k dataset\footnote{\url{https://huggingface.co/datasets/Salesforce/APIGen-MT-5k}} consisting of data points from the Retail and Airline domains. We sample 1,000 conversations and process them following the methodology proposed in Pre-Act~\cite{rawat2025preactmultistepplanningreasoning} : each conversation is decomposed into individual turns, where each turn becomes a training sample with the complete dialogue history up to that point as context. The prompt template can be found in supplementary Section \ref{sec:appendix}. This yields 5,469 training samples and 2,344 test samples. The cold-start reasoning data was generated using GPT-4o and manually verified for quality. The processed dataset is available at Dataset\footnote{\url{https://github.com/rawat-mrinal06/agentic_conversation_dataset/}}.

\noindent\textbf{Almita} We use the Almita dataset~\cite{almita2024}\footnote{\url{https://github.com/zendesk/almita-dataset}} from Zendesk, consisting of 18 customer service procedures. We apply the same preprocessing pipeline as APIGen-MT-5k, resulting in 1,383 samples. This dataset is used exclusively for evaluation (no training) to assess out-of-distribution generalization.

\subsection{Training Details}

All experiments are conducted using the TRL framework~\cite{vonwerra2022trl} with Qwen3-1.7B \cite{qwen3} as the base model. For Base SFT and Cold-Start Reasoning SFT, we apply LoRA to the query, key, value, output, gate, up, and down projection layers with rank $r=32$, $\alpha=128$, and dropout $p=0.1$. We train with a batch size of 2 and gradient accumulation steps of 4 (effective batch size of 8), using a learning rate of $2 \times 10^{-4}$ for 3 epochs. For RL, we continue training from the cold-start checkpoint with GRPO. We use a learning rate of $5 \times 10^{-6}$, batch size of 1 with gradient accumulation steps of 16 (effective batch size of 16), and generate 8 outputs per prompt for group advantage estimation. The KL coefficient $\beta$ is set to 0 and the clipping parameter $\epsilon$ to 0.2.

\begin{table*}[]
\renewcommand{\arraystretch}{1.5}
\resizebox{\textwidth}{!}{
\begin{tabular}{|l|c|c|c|c|c|c|c|c|c|c|}
\hline
\multicolumn{11}{|c|}{\textbf{APIGen-MT (Test Set)}} \\ \hline
\multirow{2}{*}{\textbf{Model}} & 
\makecell{\textbf{Action}} & 
\multicolumn{5}{c|}{\textbf{Tool}} & 
\multicolumn{4}{c|}{\textbf{Answer}} \\ \cline{3-7} \cline{8-11}
& 
\makecell{\textbf{Recall}} & 
\makecell{\textbf{Recall}} & 
\makecell{\textbf{Precision}} & 
\makecell{\textbf{F1}} & 
\makecell{\textbf{Name} \\ \textbf{Accuracy}} & 
\makecell{\textbf{Args} \\ \textbf{EM}} & 
\makecell{\textbf{Recall}} & 
\makecell{\textbf{Precision}} & 
\makecell{\textbf{F1}} & 
\makecell{\textbf{Sim.}} \\ \hline
Qwen3-1.7B  & 0.5890 & 0.6177 & 0.6084 & 0.6130 & 0.3771 & 0.5785 & 0.5847 & 0.5942 & 0.5894 & 0.7565 \\
+ SFT (no think)  & 0.8912 & \textbf{0.9463} & 0.8583 & 0.9002 & 0.8512 & 0.7234 & 0.8330 & \textbf{0.9355} & 0.8813 & 0.9731 \\
+ SFT (Cold Start - think)       & 0.8745 & 0.8906 & \textbf{0.8726} & 0.8815 & 0.8070 & 0.7098 & 0.8612 & 0.8807 & 0.8708 & 0.9650 \\
+ RL & \textbf{0.9018} & 0.9141 & 0.8993 & \textbf{0.9066} & \textbf{0.8507} & \textbf{0.7255} & \textbf{0.8903} & 0.9063 & \textbf{0.8982} & \textbf{0.9751} \\ \hline\hline
\multicolumn{11}{|c|}{\textbf{Almita (Out-of-Domain)}} \\ \hline
\multirow{2}{*}{\textbf{Model}} & 
\makecell{\textbf{Action}} & 
\multicolumn{5}{c|}{\textbf{Tool}} & 
\multicolumn{4}{c|}{\textbf{Answer}} \\ \cline{3-7} \cline{8-11}
& 
\makecell{\textbf{Recall}} & 
\makecell{\textbf{Recall}} & 
\makecell{\textbf{Precision}} & 
\makecell{\textbf{F1}} & 
\makecell{\textbf{Name} \\ \textbf{Accuracy}} & 
\makecell{\textbf{Args} \\ \textbf{EM}} & 
\makecell{\textbf{Recall}} & 
\makecell{\textbf{Precision}} & 
\makecell{\textbf{F1}} & 
\makecell{\textbf{Sim.}} \\ \hline
Qwen3-1.7B       & 0.6700 & \textbf{0.7902} & 0.4836 & 0.6000 & \textbf{0.7610} & \textbf{0.7721} & 0.6350 & 0.8750 & 0.7359 & 0.6750 \\
+ SFT (no think)  & 0.8365 & 0.7518 & 0.7192 & 0.7351 & 0.7399 & 0.7174 & 0.8724 & 0.8899 & 0.8811 & 0.7723 \\
+ SFT (Cold Start - think)       & 0.8452 & 0.7057 & \textbf{0.7642} & 0.7338 & 0.6818 & 0.7118 & \textbf{0.9056} & 0.8765 & 0.8908 & 0.7315 \\
+ RL & \textbf{0.8523} & 0.7470 & 0.7347 & \textbf{0.7408} & 0.7527 & 0.7227 & 0.8828 & \textbf{0.8992} & \textbf{0.8909} & \textbf{0.7751} \\ \hline
\end{tabular}
}
\caption{Results on APIGen-MT and Almita dataset. Models are sequentially fine-tuned: SFT (cold start - think) on SFT (no think), and RL on SFT (cold start - think).}
\label{tab:results_ind_ood}
\end{table*}
\subsection{Baselines}

We compare four model configurations to assess the incremental contribution of each training stage: (1) \textbf{Qwen3-1.7B}: the vanilla base model without any fine-tuning, serving as our baseline; (2) \textbf{Base SFT}: the model fine-tuned on task data without reasoning traces; (3) \textbf{Cold-Start SFT (with reasoning)}: the model fine-tuned with a small reasoning-annotated dataset on top of Base SFT; and (4) \textbf{RL}: the final model trained with GRPO on top of Cold-Start SFT. This progression allows us to isolate the impact of reasoning initialization and reinforcement learning.

\subsection{Evaluation Metrics}

Following Pre-Act~\cite{rawat2025preactmultistepplanningreasoning}, we evaluate model performance at each conversation turn by comparing predictions against ground truth. For each user request, the model must decide between two actions: generating a answer or making a tool call. We compute the following metrics:

\noindent\textbf{Action Classification} We measure recall for the binary action decision (tool call vs. answer). This captures the model's ability to correctly identify when external tools are needed versus when it can respond directly.

\noindent\textbf{Tool Call Quality.} When the ground truth requires a tool call, we evaluate: (1) tool name accuracy, whether the correct tool is selected; and (2) parameter match (full), whether all required parameters are provided with correct values.

\noindent\textbf{Answer Quality.} When the ground truth is a direct answer, we use a cross-encoder similarity model~\cite{xiao2024cpackpackedresourcesgeneral} to measure semantic similarity between the predicted and ground truth responses, capturing correctness beyond exact string matching.

\begin{figure*}[htbp]
    \centering
    \setlength{\abovecaptionskip}{1pt}
    \setlength{\belowcaptionskip}{0pt}
    
    \begin{subfigure}[t]{0.32\textwidth}
        \includegraphics[trim={2.3cm 0.5cm 2.5cm 1.5cm}, width=\textwidth]{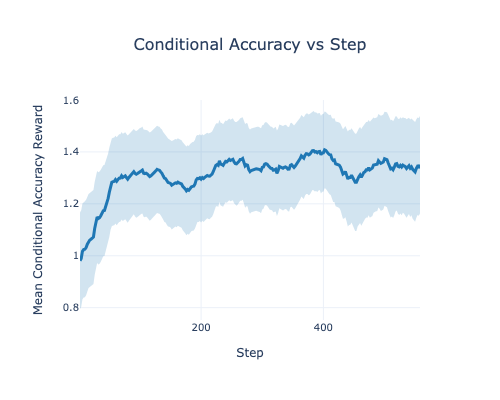}
    \end{subfigure}
    \hspace{2pt}
    \begin{subfigure}[t]{0.32\textwidth}
        \includegraphics[trim={2.3cm 0.5cm 2.5cm 2.5cm}, width=\textwidth]{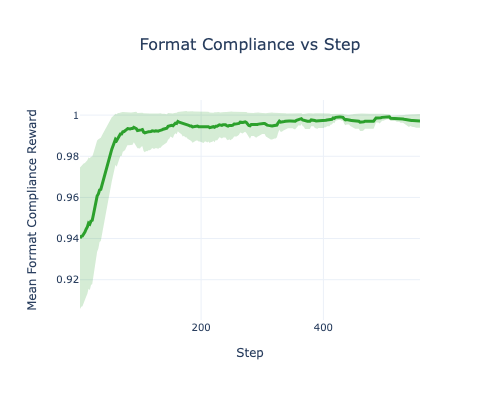}
    \end{subfigure}
    \hspace{2pt}
    \begin{subfigure}[t]{0.32\textwidth}
        \includegraphics[trim={2.3cm 0.5cm 2.5cm 1.5cm}, width=\textwidth]{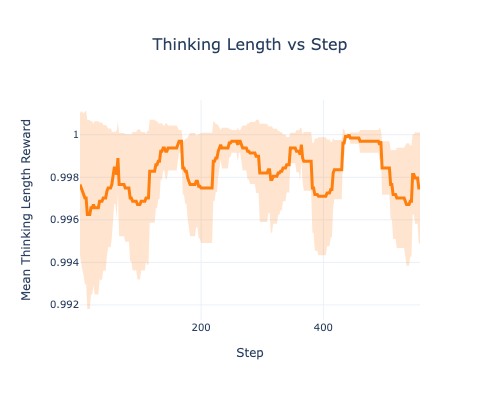}
    \end{subfigure}
    
    \caption{Reward curves for the training process.}
    \label{fig:rewards}
\end{figure*}

\begin{figure}[ht!]
\centering
     \includegraphics[trim={2.3cm 0.5cm 2.5cm 1.5cm},clip,scale=0.45]{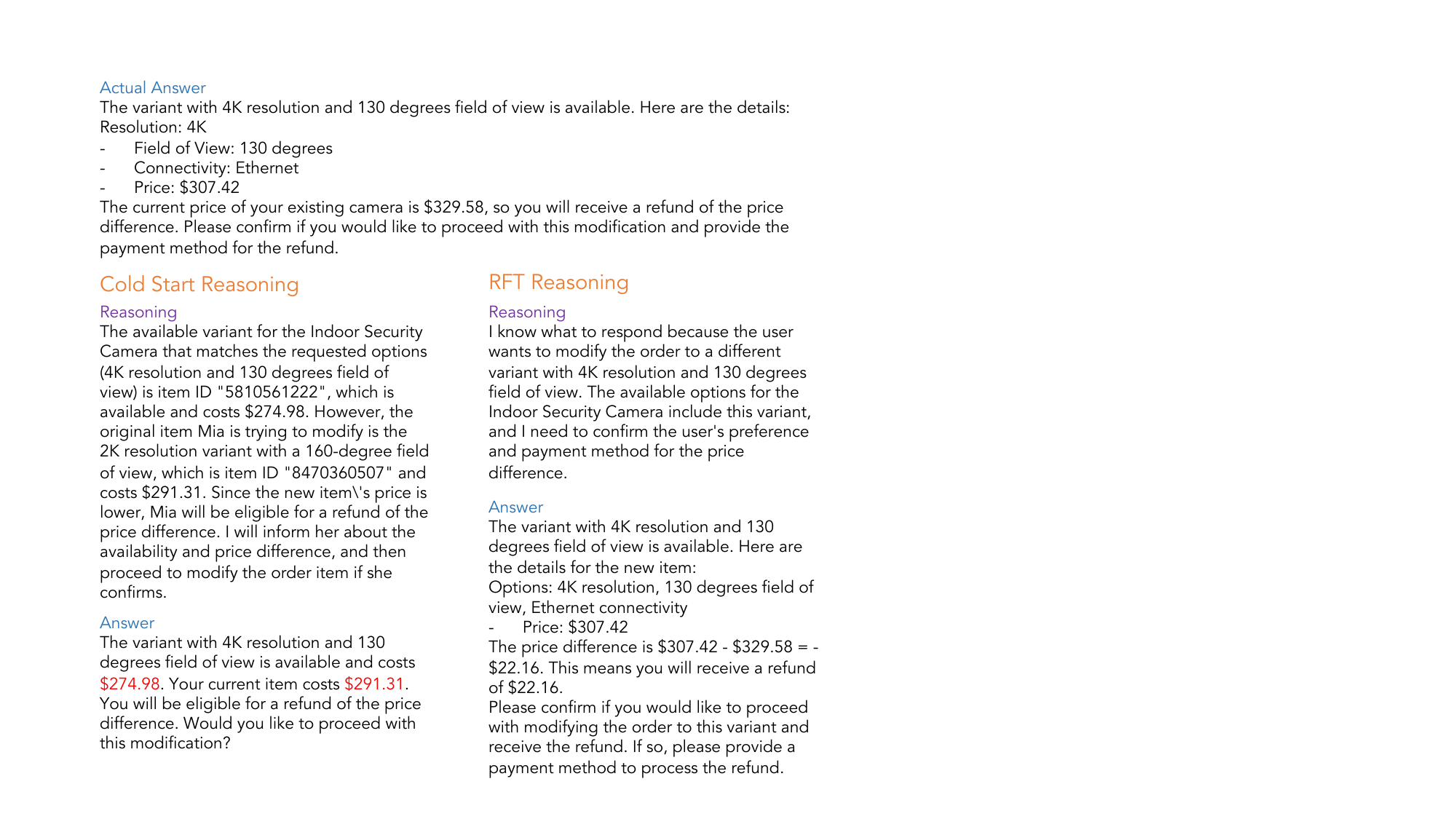}
     \caption{Example Generation: Cold-Start Model vs. RL Model}
       \label{fig:ex}

\end{figure}

\section{Results and Discussion}
Table~\ref{tab:results_ind_ood} demonstrates consistent improvements across our training pipeline on both the APIGen-MT test set and the out-of-domain Almita dataset. yields relative gains in action recall of \textbf{1.18\%} over Base SFT (no think) on APIGen-MT and \textbf{1.88\%} on Almita. Compared to the vanilla base model, we observe substantial relative improvements of \textbf{53\%} on APIGen-MT and \textbf{27.2\%} on Almita.

Base SFT without reasoning achieves high tool recall but struggles with tool selection accuracy, indicating the model learns \emph{when} to call tools but not \emph{which} tool to invoke. Cold-Start reasoning initialization shows a slight performance drop compared to Base SFT, yet significantly better than the vanilla and, importantly, retains most of the knowledge acquired in the previous stage with minimal degradation. This knowledge preservation is attributable to LoRA's parameter-efficient approach, where only a small fraction of weights are updated, preventing catastrophic forgetting. RL subsequently refines this balance, achieving the best F1 scores across both tool calls and answers. The consistent out-of-domain improvements on Almita demonstrate that reasoning-enhanced training develops transferable problem-solving strategies rather than memorizing domain-specific patterns.

\subsection{Reward Progression Analysis}

Figure~\ref{fig:rewards} reveals the optimization trajectory of our composite reward. Format compliance rapidly stabilizes within 50 steps, as expected from Cold-Start initialization. Conditional accuracy shows steady improvement from 1.0 to 1.4 over 300 steps with decreasing variance, indicating more consistent and accurate decision-making. After step 300, rewards plateau, with a slight decline suggesting potential reward model saturation. Notably, thinking length remains stable at ~0.998, showing the model maintains adequate reasoning depth without verbosity.

\subsection{Reasoning Quality Over Length}
Analysis of reasoning traces reveals a counterintuitive finding as also concluded by TooRL \cite{qian2025toolrlrewardtoollearning}: RL model generates \textbf{25\%} shorter reasoning (\textbf{60} vs \textbf{80} tokens on average) while achieving higher accuracy. Figure~\ref{fig:ex} illustrates this phenomenon. The Cold-Start model produces verbose reasoning with factual hallucinations, inventing original item specifications and miscalculating price differences (errors in \textcolor{red}{red}). In contrast, the RL model focuses only on decision-relevant information: identifying the requested variant, computing correct pricing, and confirming the action.

This suggests \textbf{RL learns to prioritize correctness over comprehensiveness}. The conditional accuracy reward penalizes incorrect outputs regardless of reasoning length, while format and length rewards prevent both insufficient and excessive verbosity. The result is focused reasoning that avoids:
\begin{itemize}[noitemsep, topsep=1pt]
    \item Hallucinating unnecessary context or facts.
    \item Performing incorrect arithmetic while appearing thorough.
    \item Including irrelevant details that may contradict the final action.
\end{itemize}

Key insights from our experiments: (1) Even ~100 reasoning samples can effectively bootstrap RL, guiding structured thinking. (2) RL improves the precision-recall trade-off, yielding more confident decisions. (3) Carefully designed composite rewards balance objectives wixthout instability. (4) Explicit reasoning enhances generalization, with out-of-domain gains indicating strategy learning over pattern matching. (5) Crucially, concise, targeted reasoning under proper constraints outperforms verbose, error-prone traces, challenging assumptions about reasoning length and quality.

\section{Conclusion and Future Work}
In this work, we propose a three-stage pipeline, in which the LLMs learn to generate reasoning steps that guide both tool usage (e.g., tool calls) and final answer generation in conversational agents. In future work, we aim to formalize the reward functions to avoid tuning constants and reduce the risk of overfitting, and to extend this approach to smaller language models (SLMs) for agentic conversational scenarios and on-edge deployment of privacy-preserving, personalized agents.

\bibliography{custom}

\appendix

\section{Prompt Template}
\label{sec:appendix}
\begin{figure*}[ht!]
\centering
     \includegraphics[trim={0.3cm 0.5cm 0.5cm 0.4cm},clip,scale=0.75]{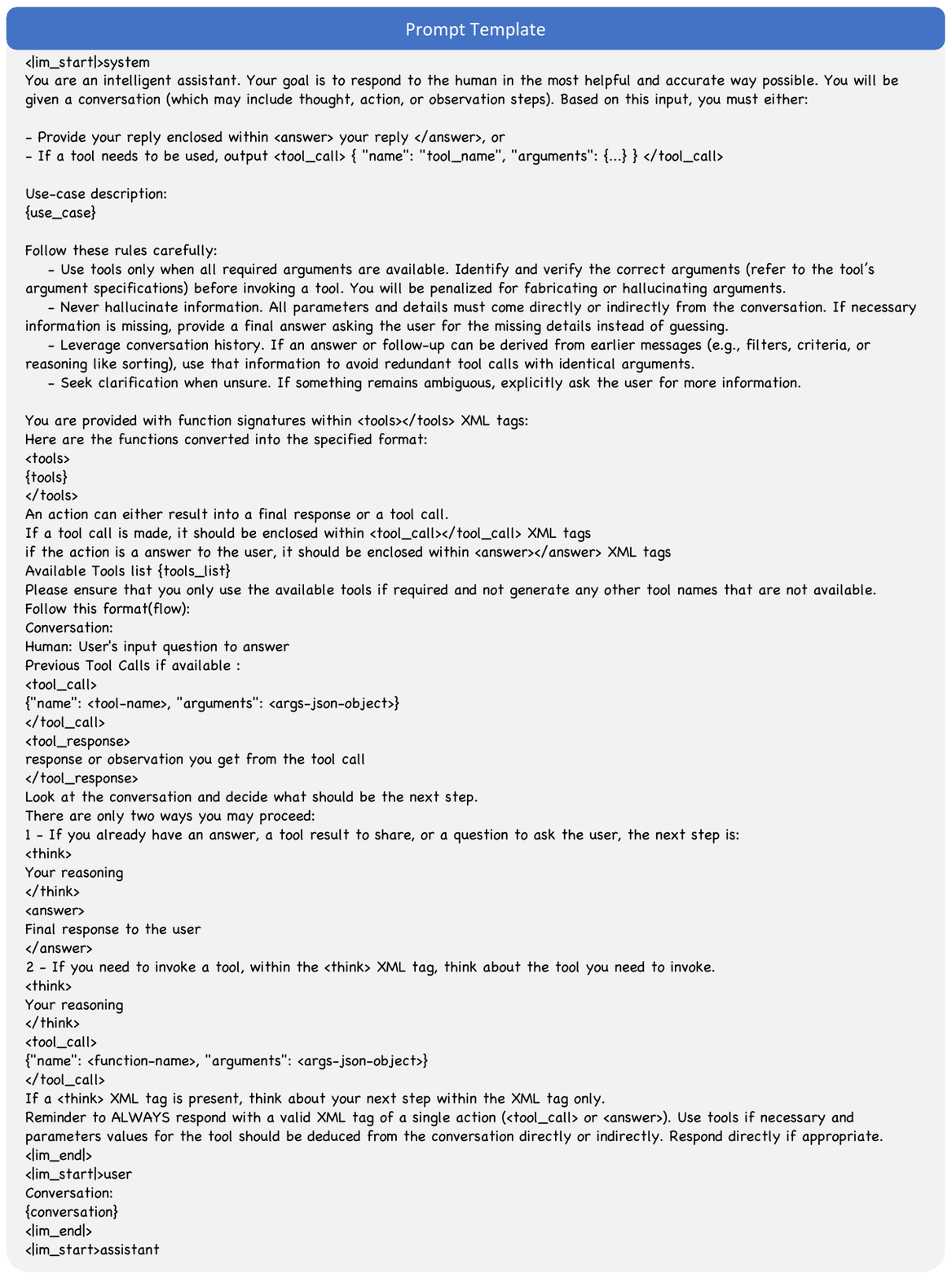}
      \caption{Prompt Template used for the dataset}
       \label{fig:prompt-e2e}

\end{figure*}

\end{document}